\documentclass{article}
\usepackage{PRIMEarxiv}

\usepackage[utf8]{inputenc}
\usepackage{adjustbox}
\usepackage[T1]{fontenc}
\usepackage{hyperref}
\usepackage{url}
\usepackage{booktabs}
\usepackage{multirow}
\usepackage{amsfonts}
\usepackage{nicefrac}
\usepackage{microtype}
\usepackage{fancyhdr}
\usepackage[table]{xcolor}

\usepackage{graphicx}
\usepackage[percent]{overpic}
\graphicspath{{media/}}
\usepackage{pgfplots}
\pgfplotsset{compat=1.18}
\usepackage{amsmath}
\usepackage{gensymb}
\usepackage{subcaption}
\usepackage{pdfpages}
\usepackage{longtable}
\usepackage{setspace}
\usepackage{cite}
\usepackage{appendix}
\usepackage{chngcntr}
\usepackage{cleveref}
\usepackage{float}
\usepackage{placeins}
\AtBeginEnvironment{appendices}{%
  \crefalias{section}{appendix}%
  \counterwithin{figure}{section}%
  \counterwithin{table}{section}%
  \counterwithin{equation}{section}%
}

\title{Two-stage Residual Learning for Modelling Strength-Informed Optimisation of Thermal Systems}

\author{
Talha Ansar \\
University of Engineering \\
\& Technology, Lahore
\And
Waqar Muhammad Ashraf \\
University College London \\
\AND
Abdulelah S. Alshehri \\
King Saud University
\And
Omar Bayomie \\
University College London \\
\texttt{omar.bayomie.23@ucl.ac.uk}
}

\author{
\begin{minipage}[t]{\textwidth}
\centering
\textbf{Talha Ansar}\textsuperscript{1} \quad
\textbf{Waqar Muhammad Ashraf}\textsuperscript{2} \quad
\textbf{Abdulelah S. Alshehri}\textsuperscript{3} \quad
\textbf{Omar Bayomie} \textsuperscript{4,5,6}$^\dagger$ \\[2mm]
\textsuperscript{1}Department of Mechanical Engineering, University of Engineering and Technology Lahore, Kala Shah Kaku, Pakistan \\
\textsuperscript{2}The Sargent Centre for Process Systems Engineering, Department of Chemical Engineering, University College London, London, UK \\[2mm]
\textsuperscript{3}Department of Chemical Engineering, College of Engineering, King Saud University, Riyadh, Saudi Arabia \\[2mm]
\textsuperscript{4}The Advanced Centre for Biochemical Engineering, Department of Biochemical Engineering, University College London, London, UK \\[2mm]
\textsuperscript{5}SSPC, The Science Foundation Ireland Research Centre for Pharmaceuticals, University College Dublin, Belfield, Dublin, Ireland \\[2mm]
\textsuperscript{6}Institut des Systemes Intelligents et de Robotique, Sorbonne University, Paris 75005, France \\[2mm]
\textsuperscript{$\dagger$}Corresponding author: \texttt{omar.bayomie.23@ucl.ac.uk}
\end{minipage}
}

\begin{document}
\maketitle
\begin{abstract}
Aggregate performance metrics do not identify local modelling inaccuracies in machine learning (ML) surrogates used for thermal-system optimisation. This study assesses local modelling strength through validation-based reliability screening and post-optimisation residual assessment. A two-stage residual learning (TSRL) framework, built on artificial neural networks (ANNs), is developed for modelling-strength-informed and domain-constrained optimisation of thermal systems. Within the TSRL framework, a base ANN predicts the process output, and a residual ANN learns the signed residuals left by the base ANN. The TSRL framework is implemented on two case studies of thermal systems, including 817 simulation observations of a crude distillation unit and operating records from a 395 MW industrial gas turbine. The TSRL framework reduces the test root mean square error (RMSE) from 10.3 g/s to 4.7 g/s for Fuel Oil in the refinery dataset and from 151 kJ/kWh to 105 kJ/kWh for turbine heat rate (THR). The base ANNs trained within TSRL are embedded in domain-constrained optimisation problems, and the modelling strength at each computed solution is assessed using the locked residual-score threshold. Each computed solution receives a good modelled region (GMR) or weak modelled region (WMR) label to inform ML model-based decision-making for safety-critical operation of thermal systems.
\end{abstract}

\keywords{surrogate modelling \and residual correction \and reliability screening \and process optimisation \and thermal systems}

\section*{Nomenclature}

\begin{tabular}{@{}ll@{}}
\toprule
\textbf{Symbol or acronym} & \textbf{Meaning} \\
\midrule
AGO & Atmospheric gas oil \\
AH & Ambient humidity \\
ANN & Artificial neural network \\
AP & Ambient pressure \\
AT & Ambient temperature \\
AUC & Area under the ROC curve \\
CDP & Compressor discharge pressure \\
CDT & Compressor discharge temperature \\
$D_M$ & Mahalanobis distance \\
$b$ & Half-range used to truncate indicator-band perturbations \\
$\delta$ & Mahalanobis-distance limit \\
$e_i$ & Absolute base-model error for observation $i$ \\
$\epsilon_k$ & $k$th sampled indicator-band perturbation \\
FGT & Fuel gas temperature at the combustion-chamber inlet \\
GFFR & Gas fuel flow rate \\
GMR & Good modelled region \\
$\mathcal{L}_B,\mathcal{L}_R$ & Base- and residual-ANN training losses \\
MAE & Mean absolute error \\
MLP & Multilayer perceptron \\
MSE & Mean squared error \\
$N$ & Number of observations in the relevant data split \\
$N_{\mathrm{ref}}$ & Number of refinery modelling observations (817) \\
PHGOT & Performance-heater gas outlet temperature \\
$r_i$ & Signed base-model residual for observation $i$ \\
RMSE & Root mean square error \\
ROC & Receiver operating characteristic \\
$s_i$ & Residual score (predicted residual magnitude) for observation $i$ \\
$\sigma_R$ & Normal scale used for indicator-band sampling \\
SiLU & Sigmoid linear unit \\
TE & Thermal efficiency \\
THR & Turbine heat rate \\
TSRL & Two-stage residual learning \\
$T_e$ & Observed-error threshold \\
$T_s$ & Residual-score threshold \\
WMR & Weak modelled region \\
$x,\tilde{x}$ & Scaled and physical-scale process-input vectors \\
$w_j,w_{ij}$ & Crude blend fractions \\
$z_{ij}$ & Raw Latin hypercube design coordinate \\
$y,\hat{y},\bar{y}$ & Observed output, model prediction, and split mean \\
$y_{\mathrm{CO_2}}^{\mathrm{cap}}$ & CO$_2$ ceiling in refinery optimisation \\
$x_j^{\mathrm{target}}$ & Prescribed scaled input target \\
$\hat{y}^{B}$ & Base-model prediction \\
$\hat{y}^{R}$ & Residual-model prediction \\
$\hat{y}^{C}$ & Residual-corrected prediction \\
$\theta_B,\theta_R$ & Trainable parameter vectors of the base and residual ANN \\
\bottomrule
\end{tabular}

\section{Introduction}

Optimising the operation of thermal systems determines operating decisions that improve a defined objective while satisfying equipment, material, energy, and safety constraints \cite{jaluria2019design}. Surrogate-based optimisation \cite{Forrester2009,Bhosekar2018} reduces the computational cost of model-based decision-making by replacing repeated calls to a detailed model with an approximation that can be evaluated quickly. In refinery-wide energy optimisation, artificial neural networks (ANNs) replace high-fidelity first-principles process models \cite{gueddar2012novel}. ANN-based models are also used for the data-driven design of cooling systems \cite{JamilAshraf2024}.

An ANN is a nonlinear continuous function approximator supported by universal approximation theory \cite{Hornik1989}. Its differentiable form suits surrogate-based optimisation, including thermal-power studies that use machine-intelligence models in constrained operation \cite{AshrafDua2024}. The explicit algebraic form of a trained ANN can be embedded in a mathematical programme. Deterministic global optimisation \cite{SchweidtmannMitsos2019}, the OMLT toolkit \cite{Ceccon2022OMLT}, and a recent gas-turbine simulation platform \cite{AshrafOperator2026} evaluate trained networks inside constrained optimisation. Misener and Biegler \cite{MisenerBiegler2023} identify robustness to uncertainty and accurate extrapolation as essential conditions for surrogates used in process optimisation. Parameter estimation, model selection, and final evaluation require separate data to avoid optimistic error estimates from reusing selection data \cite{XuGoodacre2018,VarmaSimon2006}. Root mean square error (RMSE), mean absolute error (MAE), and the coefficient of determination ($R^2$) summarise split-level performance, but do not locate modelling errors within the learned input--output mapping.

Limited information about local modelling accuracy at given input conditions restricts the use of surrogate-based optimisation in the operational control of thermal systems. The optimiser searches for favourable predictions and may return a point with high local modelling error. Kahrs and Marquardt \cite{KahrsMarquardt2007} bounded the optimisation of hybrid models by a validity domain built from the training inputs. Schweidtmann et al. \cite{Schweidtmann2022Obey} combined topological data analysis with one-class support vector machines and convex hulls, and encoded the data domain as optimisation constraints. Meyer and Pebesma \cite{MeyerPebesma2021} introduced an area of applicability for spatial prediction models with a cross-validation-derived threshold to identify regions where predictive errors are expected to be comparable to cross-validation errors. Recent gas-turbine studies \cite{AshrafDomain2026,AshrafOperator2026} use a Mahalanobis constraint to retain the multivariate input combinations represented in plant data. These methods restrict surrogate evaluation to observed input combinations, but satisfying a domain constraint does not establish local prediction accuracy.

Domain restrictions can be complemented by predictive uncertainty estimates from Monte Carlo dropout \cite{GalGhahramani2016} and deep ensembles \cite{Lakshminarayanan2017}. Gawlikowski et al. \cite{Gawlikowski2023} reviewed their assumptions, computational cost, and performance. Cajahuaringa et al. \cite{Cajahuaringa2024} applied Monte Carlo dropout to a nonlinear autoregressive model of a 215 MW gas turbine and reported the uncertainty of the predicted rotational speed. The study addressed predictive uncertainty but did not carry the uncertainty into surrogate-based operational optimisation of the gas turbine system. Yang et al. \cite{YangLiWu2021} trained an ensemble of ReLU networks on resampled training sets, approximated the ensemble with a mixture density network, and solved a chance-constrained mixed-integer programme. These approaches require repeated training or stochastic forward passes. When used in optimisation, the uncertainty model enters the problem and changes its size and form. The resulting spread measures disagreement among networks or samples, rather than a decision threshold calibrated against observed errors of the deployed surrogate.

Residual learning offers a direct connection to observed errors. In stagewise function fitting \cite{Friedman2001}, each new learner is fitted to the errors left by the preceding model. Computer-model calibration \cite{KennedyOHagan2001} treats model discrepancy as a term separate from the simulator response. Hybrid semi-parametric models \cite{vonStosch2014,Bradley2022} use a data-driven component to correct a mechanistic model. In these studies, the residual learner improves prediction accuracy. The size of the learned correction is not used as a reliability signal. Lahlou et al. \cite{Lahlou2023DEUP} trained a secondary predictor of the pointwise error of a main model as a measure of epistemic uncertainty. Their method was developed for active learning, sequential model optimisation, and exploration in reinforcement learning. It returns a continuous uncertainty estimate. It does not lock a decision threshold on held-out data, and it does not label the solutions of a constrained nonlinear programme.

This work calibrates a residual score against observed base-model errors on a validation split, then assesses solutions of domain-constrained optimisation without adding an uncertainty model to its formulation. Refinery and gas-turbine surrogates are embedded in optimisation studies \cite{gueddar2012novel,AshrafDomain2026,AshrafOperator2026}, yet the reported operating points do not carry a point-specific measure of modelling strength before they reach the operator.

This work develops a two-stage residual learning (TSRL) framework for assessing local modelling strength in surrogate-based optimisation. In the first stage, a base ANN is fitted for each process output. In the second stage, the base parameters are fixed and a residual ANN learns the signed base-model error. The signed residual gives a point correction. Its magnitude gives a residual score, which remains available at an operating point where the true output is unknown. Validation data link the score to the observed base error and determine two percentile thresholds by maximising $F_1$. The thresholds are locked before test evaluation and before any optimisation solution is scored. Only the base ANN is embedded in the optimisation problem. The residual score then labels each returned solution as lying in a good modelled region (GMR) or a weak modelled region (WMR).

The TSRL framework is examined on two thermal systems with different sources of modelling error. The first case uses 817 simulated observations of a crude distillation unit, with seven inputs that cover varied crude blends and utility conditions. The second case uses operating records of a 395 MW industrial gas turbine \cite{AshrafDua2024}, where the measurements carry variability from operating dynamics and sensor noise. Gas-turbine studies \cite{liu2020gas,AshrafDomain2026,AshrafOperator2026} report ML-based prediction and optimisation of plant performance without a point-specific check of modelling strength at new input conditions. TSRL adds local reliability assessment to surrogate-based optimisation through three main contributions.

\begin{itemize}
\item The TSRL framework uses a residual ANN to correct the base surrogate and its predicted residual magnitude to score reliability at unseen operating points.

\item Residual scores are calibrated by cross-matching percentile labels of observed base error and residual score for each validation observation. The threshold pair is selected by $F_1$, and both thresholds are locked before testing.

\item Computed solutions are assessed after optimisation while retaining the original optimisation problem. Base ANNs are embedded in Mahalanobis-constrained nonlinear programmes, and each solution receives a GMR or WMR label from TSRL after convergence. Many solutions of the refinery case study satisfy the domain constraints and yet fall in the WMR.

\end{itemize}

The rest of the paper is organised as follows. Section~\ref{sec_methods} defines the models, calibration procedure, and optimisation formulation. Section~\ref{sec_results} presents the refinery and gas turbine studies. Section~\ref{sec_conclusion} summarises the findings, limitations, and directions for further work.

\section{Methods}
\label{sec_methods}

\begin{figure}[t]
  \centering
  \includegraphics[width=0.94\linewidth]{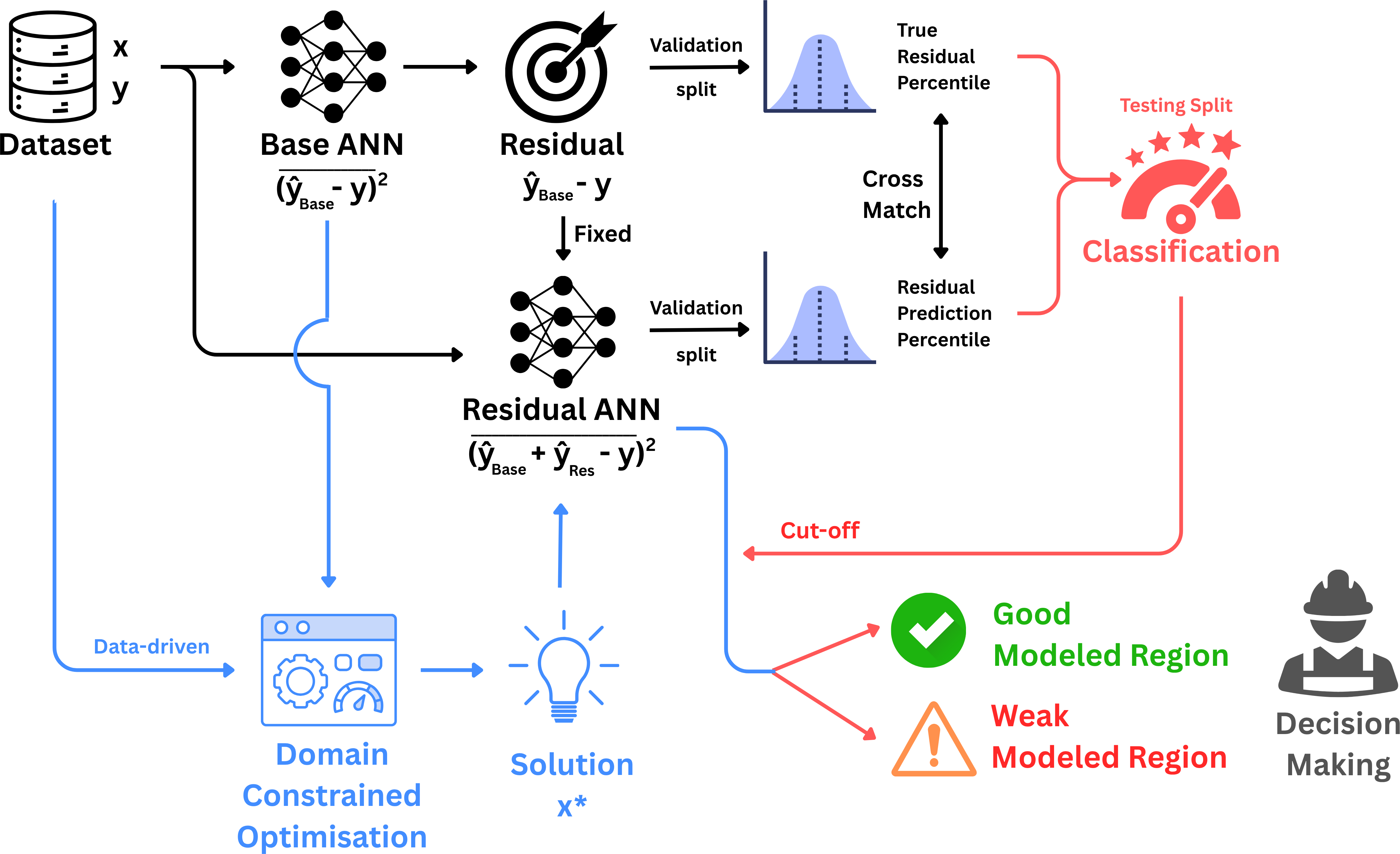}
  \caption{TSRL information flow. The data contain process inputs $x$ and observed output $y$, while the base and residual ANNs predict $\hat{y}^{B}$ and $\hat{y}^{R}$, respectively. The residual target is $r_i=y_i-\hat{y}^{B}_i$. Validation labels based on absolute base error and residual score fix $T_e$ and $T_s$ before testing. Only the base ANN enters the optimisation problem, and $|\hat{y}^{R}(x^*)|$ assigns a GMR or WMR label after convergence.}
  \label{fig_methodology}
\end{figure}

The TSRL framework separates surrogate fitting, residual correction, reliability calibration, and optimisation, as illustrated in Figure~\ref{fig_methodology}. For each process output, a base ANN is fitted and its prediction, $\hat{y}^{B}(x)$, is used as the surrogate response in the optimisation problem. After the base-network parameters are fixed, a residual ANN receives the same input vector $x$ and predicts the signed base-model error, $\hat{y}^{R}(x)$. The corrected response is $\hat{y}^{C}(x)=\hat{y}^{B}(x)+\hat{y}^{R}(x)$, while the magnitude $|\hat{y}^{R}(x)|$ is retained separately as a residual score. This score remains available at a proposed operating point where the true process output is unknown.

Each validation observation has an observed absolute base error and a residual score. A percentile search applies an error threshold to the first quantity and a score threshold to the second quantity. The two resulting labels are cross-matched sample by sample to calculate validation precision, recall, and $F_1$ for each threshold pair. The pair that maximises validation $F_1$ fixes the observed-error threshold $T_e$ and residual-score threshold $T_s$. Both thresholds are locked before test evaluation \cite{XuGoodacre2018,VarmaSimon2006}. The domain-constrained optimisation then embeds only the base ANN and returns an operating solution $x^*$. The residual score at $x^*$ is compared with the locked $T_s$ to assign a GMR or WMR label.

\subsection{Data partition, scaling, and evaluation}

Each dataset of the two case studies is divided randomly into 70\% training observations, 15\% validation observations, and 15\% test observations. The training split is used to estimate network parameters. The validation split guides hyperparameter selection, early stopping, output selection for detailed reliability analysis, and threshold calibration. The test split is evaluated once after modelling choices have been fixed. Each input is normalised to the interval $[0,1]$ by min-max scaling,

\begin{equation}
x_{ij}=
\frac{\tilde{x}_{ij}-\tilde{x}_{j}^{\min}}
{\tilde{x}_{j}^{\max}-\tilde{x}_{j}^{\min}},
\label{eq_scaling}
\end{equation}

where $i$ indexes an observation and $j$ indexes an input variable. $\tilde{x}_{ij}$ is the physical-scale value of input $j$ at observation $i$, while $\tilde{x}_{j}^{\min}$ and $\tilde{x}_{j}^{\max}$ are the lower and upper values used to scale that input. The resulting $x_{ij}$ is dimensionless and lies in $[0,1]$. The same limits are used for all splits and to convert an optimisation solution between scaled and physical input values.

The modelling performance of each ANN is evaluated using RMSE, MAE, and $R^2$,

\begin{equation}
\begin{aligned}
\mathrm{RMSE}
&=
\sqrt{\frac{1}{N}\sum_{i=1}^{N}
\left(y_i-\hat{y}_i\right)^2},\\
\mathrm{MAE}
&=
\frac{1}{N}\sum_{i=1}^{N}
\left|y_i-\hat{y}_i\right|,\\
R^2
&=
1-
\frac{\sum_{i=1}^{N}\left(y_i-\hat{y}_i\right)^2}
{\sum_{i=1}^{N}\left(y_i-\bar{y}\right)^2}.
\end{aligned}
\label{eq_regression_metrics}
\end{equation}

For the evaluated split, $N$ is the number of observations and $\bar{y}$ is its mean output, while $\hat{y}_i$ denotes the model prediction being assessed.

\subsection{Base and residual surrogate models}

The base and residual ANNs use a multilayer perceptron (MLP) architecture for nonlinear input--output modelling \cite{Hornik1989,AshrafDINN2024,liu2020gas}. A separate multi-input single-output network is fitted for every process output. Each network has one hidden layer and a linear output. The hidden units use the sigmoid linear unit (SiLU) activation \cite{Elfwing2018}, with $\operatorname{SiLU}(u)=u/(1+\exp(-u))$. The base surrogate is written as

\begin{equation}
\hat{y}^{B}
=
W_{\mathrm{out}}
\operatorname{SiLU}\left(W_{\mathrm{in}}x+b_{\mathrm{in}}\right)
+b_{\mathrm{out}}.
\label{eq_base}
\end{equation}

$W_{\mathrm{in}}$ and $W_{\mathrm{out}}$ contain the input-to-hidden and hidden-to-output network weights, while $b_{\mathrm{in}}$ and $b_{\mathrm{out}}$ contain the corresponding bias vectors. This differentiable expression can be embedded in a nonlinear programme \cite{SchweidtmannMitsos2019,AshrafOperator2026}. The base-ANN loss combines mean squared error (MSE) with an L1 penalty,

\begin{equation}
\mathcal{L}_{B}
=
\frac{1}{N}\sum_{i=1}^{N}
\left(y_i-\hat{y}^{B}_i\right)^2
+\lambda_{1,B}\lVert\theta_B\rVert_1.
\label{eq_base_loss}
\end{equation}

Regularisation is applied to the base-network parameter vector $\theta_B$. The coefficient $\lambda_{1,B}$ weights the L1 norm $\lVert\theta_B\rVert_1$. The L2 coefficient ($\lambda_{2,B}$) is applied through Adam weight decay. The learning rate, L1 coefficient, L2 coefficient, and hidden-layer size are tuned in 50 trials for each network using Hyperopt \cite{Bergstra2015} with the tree-structured Parzen estimator \cite{Bergstra2011} over the search space in Table~\ref{tab_hyperparameters}. The candidate with the lowest validation RMSE is retained.

\begin{table}[H]
\centering
\caption{Hyperparameter search space used independently for the base and residual ANNs.}
\label{tab_hyperparameters}
\begin{tabular}{lcc}
\toprule
\textbf{Hyperparameter} & \textbf{Search range} & \textbf{Sampling type} \\
\midrule
Learning rate & $10^{-5}$ to $10^{-1}$ & Log continuous \\
L1 coefficient & $10^{-7}$ to $10^{-1}$ & Log continuous \\
L2 coefficient & $10^{-7}$ to $10^{-1}$ & Log continuous \\
Hidden-layer neurons & 2 to 32 & Discrete in steps of 2 \\
\bottomrule
\end{tabular}
\end{table}

Training uses the Adam optimiser \cite{Kingma2014} in PyTorch \cite{Paszke2019} with mini-batch gradients, a maximum of 5000 epochs, and validation-based early stopping after 200 successive epochs without improvement in RMSE \cite{Prechelt1998}.

After fitting the base ANN, $\theta_B$ is fixed and the signed residual target for the second stage is computed as \cite{Friedman2001}

\begin{equation}
r_i=y_i-\hat{y}^{B}_i,
\label{eq_residual_target}
\end{equation}

where a positive $r_i$ indicates underprediction by the base ANN. A second network uses the same process inputs to predict $r_i$. Its hyperparameters are selected independently from the same ranges. The residual prediction and corrected response are

\begin{equation}
\hat{y}^{R}_i=F_R(x_i),
\qquad
\hat{y}^{C}_i=\hat{y}^{B}_i+\hat{y}^{R}_i.
\label{eq_residual_model}
\end{equation}

Here $F_R(\cdot)$ denotes the residual ANN mapping. The residual loss is expressed as

\begin{equation}
\mathcal{L}_{R}
=
\frac{1}{N}\sum_{i=1}^{N}
\left(r_i-\hat{y}^{R}_i\right)^2
+\lambda_{1,R}\lVert\theta_R\rVert_1.
\label{eq_residual_loss}
\end{equation}

The residual-network parameter vector is $\theta_R$, and $\lambda_{1,R}$ weights its L1 penalty. L2 weight decay $\lambda_{2,R}$ is applied during training. The corrected prediction is evaluated without updating the base network.

For each case study, one output is carried forward for detailed residual-score assessment. The selection uses the relative validation improvement,

\begin{equation}
\Delta_{\mathrm{val}}
=
\frac{
\mathrm{RMSE}^{B}_{\mathrm{val}}
-
\mathrm{RMSE}^{C}_{\mathrm{val}}
}{
\mathrm{RMSE}^{B}_{\mathrm{val}}
},
\label{eq_validation_improvement}
\end{equation}

where $\mathrm{RMSE}^{B}_{\mathrm{val}}$ and $\mathrm{RMSE}^{C}_{\mathrm{val}}$ are the validation RMSE values of the base and corrected predictions, respectively. A positive $\Delta_{\mathrm{val}}$ means that residual correction lowers validation RMSE. The selected output has the largest positive $\Delta_{\mathrm{val}}$ among the outputs for which training RMSE also decreases. This rule is applied before any test result is examined.

\subsection{Residual-score calibration and reliability classification}

The validation split supplies the empirical link between observed base-model error and the residual score. For each validation observation,

\begin{equation}
e_i=\left|y_i-\hat{y}^{B}_i\right|,
\qquad
s_i=\left|\hat{y}^{R}_i\right|.
\label{eq_scores}
\end{equation}

The observed absolute base error $e_i$ requires a known output $y_i$, while the residual score $s_i$ is also available at a new operating point. Both quantities are calculated for the same validation observation before threshold calibration. Spearman rank correlation is calculated between $e_i$ and $s_i$.

The continuous quantities are converted to reference and predicted classes with percentile thresholds,

\begin{equation}
\begin{aligned}
T_e(p_e)&=P_{p_e}(e),
&
z_i^{\mathrm{true}}
&=
\mathbb{I}\left[e_i\geq T_e(p_e)\right],\\
T_s(p_s)&=P_{p_s}(s),
&
z_i^{\mathrm{pred}}
&=
\mathbb{I}\left[s_i\geq T_s(p_s)\right].
\end{aligned}
\label{eq_thresholds}
\end{equation}

Here, $P_p(\cdot)$ denotes the empirical $p$th percentile, $p_e$ and $p_s$ are the candidate error and score percentiles, and $T_e$ and $T_s$ are the resulting numerical thresholds. $\mathbb{I}[\cdot]$ is an indicator function that equals one when its condition is met and zero otherwise. Thus, $z_i^{\mathrm{true}}=1$ identifies a validation observation with high observed base error, while $z_i^{\mathrm{pred}}=1$ identifies an observation with a high residual score. The candidate grids are $p_e,p_s\in\{80,81,\ldots,95\}$. For every pair, these two labels are cross-matched for the same validation observations, and precision, recall, and $F_1$ are computed as

\begin{equation}
\mathrm{Precision}
=
\frac{\mathrm{TP}}{\mathrm{TP}+\mathrm{FP}},
\qquad
\mathrm{Recall}
=
\frac{\mathrm{TP}}{\mathrm{TP}+\mathrm{FN}},
\qquad
F_1
=
\frac{2\,\mathrm{Precision}\,\mathrm{Recall}}
{\mathrm{Precision}+\mathrm{Recall}}.
\label{eq_classification_metrics}
\end{equation}

In these expressions, TP is the number of observations labelled high by both $z_i^{\mathrm{true}}$ and $z_i^{\mathrm{pred}}$, FP is the number labelled high by the score but not by the observed error, and FN is the number labelled high by the observed error but not by the score. The pair with the highest validation $F_1$ is selected. Receiver operating characteristic (ROC) curves and their area under the curve (AUC) are calculated from the validation labels and residual scores.

The selected $T_e$ and $T_s$ are fixed before test evaluation. A point with $s(x)<T_s$ is assigned to GMR. A point with $s(x)\geq T_s$ is assigned to WMR. GMR and WMR locate a point's residual score relative to the locked validation threshold.

The high-error reference class is also defined from validation percentiles, so $F_1$ measures agreement with a relative error label. Where an accepted process-error limit is available, it can replace the percentile definition of $T_e$ within the same calibration procedure.

\subsection{Constrained optimisation and local assessment}

Pyomo \cite{Bynum2021} is used to write the fitted base networks, process constraints, and objective as one nonlinear programme. The trained surrogate is evaluated within the optimisation problem \cite{SchweidtmannMitsos2019,AshrafOperator2026}. The general problem is formulated as

\begin{equation}
\begin{aligned}
\max_x\quad &J(x)\\
\text{subject to}\quad
&y_q^{\mathrm{LB}}\leq\hat{y}^{B}_q(x)\leq y_q^{\mathrm{UB}}
&&q\in\mathcal{P}\\
&D_M^2(\tilde{x})\leq\delta^2\\
&0\leq x_j\leq1
&&j\in\mathcal{X},
\end{aligned}
\label{eq_general_optimisation}
\end{equation}

where $J(x)$ is the case-specific objective and $\mathcal{P}$ is the set of constrained outputs indexed by $q$. $\hat{y}^{B}_q(x)$ is the base-ANN prediction for output $q$, and $y_q^{\mathrm{LB}}$ and $y_q^{\mathrm{UB}}$ are its lower and upper permitted values. $\mathcal{X}$ indexes the process inputs. Mahalanobis distance \cite{Ghorbani2019} accounts for covariance when measuring separation from a multivariate reference distribution. Recent gas-turbine formulations \cite{AshrafDomain2026,AshrafOperator2026} use this distance to restrict data-driven optimisation to correlated plant-input combinations. The squared distance is written as

\begin{equation}
D_M^2(\tilde{x})
=
\left(\tilde{x}-\mu\right)^{\mathsf{T}}
\Sigma^{+}
\left(\tilde{x}-\mu\right).
\label{eq_mahalanobis}
\end{equation}

The mean vector $\mu$ and covariance matrix $\Sigma$ are calculated from the reference inputs. The Moore-Penrose pseudoinverse \cite{Penrose1955}, denoted by $\Sigma^{+}$, is used when the covariance matrix is singular. The empirical limit $\delta$ restricts the search to a covariance-aware input region represented by the data. This excludes input combinations that satisfy individual bounds but are poorly represented jointly \cite{ratz2024identifying,MeyerPebesma2021}.

The nonlinear programme is solved with IPOPT through the GAMS solver interface. IPOPT implements an interior-point filter line-search method \cite{WachterBiegler2006} and returns a locally optimal constrained solution. The base surrogate remains in the objective and constraints so that residual assessment does not change the optimisation problem. After a solution $x^*$ has been returned, the residual network is evaluated at $x^*$. Its score is $s(x^*)=|\hat{y}^{R}(x^*)|$ and is compared with the fixed $T_s$.

A residual-informed indicator band is centred on the corrected prediction. Its scale is defined as

\begin{equation}
b=\frac{\left|\hat{y}^{R}(x^*)\right|}{2},
\qquad
\sigma_R=\frac{b}{1.96}.
\label{eq_interval_scale}
\end{equation}

The half-range $b$ equals half the residual score and bounds the sampled perturbations. $\sigma_R$ is the standard deviation of the normal distribution before truncation. Five hundred values are sampled from the zero-centred normal distribution and restricted to $[-b,b]$,

\begin{equation}
\epsilon_k\sim\mathcal{N}(0,\sigma_R^2),
\qquad
-b\leq\epsilon_k\leq b,
\qquad
\hat{y}^{(k)}=\hat{y}^{C}(x^*)+\epsilon_k,
\qquad
k=1,\ldots,500,
\label{eq_interval_sampling}
\end{equation}

The 2.5th and 97.5th empirical percentiles of $\hat{y}^{(k)}$ form the reported band. The indicator band visualises sampled perturbations around the corrected prediction for comparison among solutions, with its width set by the residual score.

\section{Results and Discussion}
\label{sec_results}

Model performance and residual-score screening are evaluated before computed optimisation solutions are labelled GMR or WMR. The refinery simulation is considered first because its controlled input variation provides a direct test of approximation error. The gas turbine study then applies the TSRL framework to sensor-based industrial records that contain measurement variation and a less uniform operating history.

\subsection{Case Study 1. Oil Refinery}

\subsubsection{Rigorous process model}
The refinery case study is built on a steady-state Aspen HYSYS model of an industrial crude distillation unit (CDU) that replicates an operating unit with a design throughput of 291.7~t/h. The flowsheet comprises the crude preheat train, the fired heater, and the atmospheric column with its side strippers and pump-around circuits, and delivers the product slate of the real unit, including fuel gas, LPG, stabilised naphtha, heavy naphtha, Kerosene, gas oil, and atmospheric residue (fuel oil). The four crude oils are characterised in the HYSYS Oil Manager from their laboratory assays (true-boiling-point curves, light-ends compositions, and gravity data), and vapour--liquid equilibrium is described using an equation of state (EOS). The model was validated against plant operating data in earlier work \cite{BAYOMIE2019}. The flow rates and product draw temperatures match plant records to within a few percent, and the simulated ASTM~D86 5--95\% gaps between adjacent cuts reproduce the measured separation quality. Throughout this work, this validated model is treated as the ground-truth response function whose input--output behaviour is learned by TSRL, with local modelling strength assessed for the resulting surrogates.

\subsubsection{Crude blend input space and sampling design}

The operational degrees of freedom are the blend fractions of four commercially traded crudes, namely Basrah, Arab Heavy, Arab Medium, and Kuwait, spanning the light-to-heavy API range processed by the unit. The total feed rate is held at the design throughput, so a blend is fully specified by the fraction vector $\mathbf{w} = (w_1,\dots,w_4)$ with $w_j \geq 0$ and $\sum_{j=1}^{4} w_j = 1$. Design points are generated by Latin hypercube sampling (LHS), selected for its stratified marginal coverage at moderate sample sizes \cite{McKay1979,Donovan2018}. A raw design in the four-dimensional unit hypercube was drawn with the \texttt{scipy.stats.qmc} implementation under a fixed random seed for reproducibility, and each point was projected onto the simplex by normalisation,
\begin{equation}
  w_{ij} \;=\; \frac{z_{ij}}{\sum_{\ell=1}^{4} z_{i\ell}},
  \qquad j = 1,\dots,4 .
  \label{eq:simplexnorm}
\end{equation}
The projection does not preserve the strict one-dimensional stratification of the hypercube design, and the normalised fractions concentrate mildly towards the simplex centroid relative to a uniform (flat Dirichlet) distribution. However, it retains the space-filling character of LHS over the feasible blend space while guaranteeing that every sample satisfies the unit-sum constraint exactly. No crude was constrained away from zero, allowing near-vertex blends (one crude dominant) and near-centroid blends.

\subsubsection{Automated simulation campaign}
\label{sec:datagen:automation}

The sampling campaign was executed through a programmatic Python--HYSYS interface using an object-model approach to integrate Python with the process simulator \cite{Kumar2025}. The simulator is driven over the ActiveX/COM automation layer (\texttt{pywin32}), and a HYSYS Spreadsheet operation embedded in the flowsheet serves as the single input--output interface. Its input cells are linked to the four blend fractions and its output cells import the monitored flowsheet variables, so the Python driver does not address individual unit operations directly. This design decouples the sampling logic from the flowsheet topology and makes the driver reusable across model revisions.

For each design point the driver executes a hold--write--release cycle. The solver is paused (\texttt{Solver.CanSolve = False}), the four blend fractions are written to the spreadsheet input cells, and the solver is released, after which the flowsheet re-converges from the previous solution as a warm start. The output cells are then read back before the driver proceeds to the next sample.

\subsubsection{Data description}

The refinery modelling dataset contains $N_{\mathrm{ref}}=817$ simulation observations. The seven inputs are the Basrah, Arab Heavy, Arab Medium, and Kuwait crude flow rates, reboiler duty, and two condenser loads. The seven outputs are Light Naphtha, Gases, Heavy Naphtha, Kerosene, AGO, Fuel Oil, and carbon dioxide (CO$_2$). Table~\ref{tab_refinery_statistics} reports the retained ranges. The crude rates span near-zero values to about 56--65 kg/s, which permits wide blend variation. The full reboiler-duty range, in contrast, is about 3.5\% of its mean. The gas stream is zero or near zero for much of the dataset and has a small number of larger observations. This imbalance affects its regression metrics.

\begin{table}[H]
\centering
\caption{Descriptive statistics for the refinery variables. Values use two decimal places except the low-flow gas stream, which uses four.}
\label{tab_refinery_statistics}
\scriptsize
\begin{tabular}{lccccc}
\toprule
\textbf{Variable} & \textbf{Mean} & \textbf{SD} & \textbf{Minimum} & \textbf{Median} & \textbf{Maximum} \\
\midrule
Basrah (kg/s) & 20.24 & 11.02 & 0.01 & 20.10 & 55.71 \\
Arab Heavy (kg/s) & 19.87 & 9.92 & 0.12 & 20.31 & 58.18 \\
Arab Medium (kg/s) & 20.50 & 10.98 & 0.06 & 20.76 & 58.33 \\
Kuwait (kg/s) & 20.41 & 10.84 & 0.04 & 21.09 & 64.82 \\
Reboiler duty (kW) & 35,297.00 & 193.00 & 34,638.00 & 35,300.00 & 35,860.00 \\
Condenser 1 (kW) & 6,399.96 & 239.55 & 4,850.55 & 6,417.13 & 7,021.43 \\
Condenser 2 (kW) & 13,314.49 & 439.53 & 10,230.67 & 13,316.29 & 14,629.20 \\
Light Naphtha (kg/s) & 5.24 & 0.14 & 4.80 & 5.23 & 5.63 \\
Gases (kg/s) & 0.0031 & 0.0179 & 0.0000 & 0.0000 & 0.3245 \\
Heavy Naphtha (kg/s) & 2.92 & 0.16 & 2.44 & 2.91 & 3.40 \\
Kerosene (kg/s) & 12.25 & 0.30 & 11.51 & 12.25 & 13.26 \\
AGO (kg/s) & 11.70 & 0.15 & 11.26 & 11.71 & 12.13 \\
Fuel Oil (kg/s) & 46.16 & 0.58 & 44.20 & 46.16 & 47.65 \\
CO$_2$ (kg/h) & 8,354.88 & 45.71 & 8,198.81 & 8,355.57 & 8,488.15 \\
\bottomrule
\end{tabular}
\end{table}

\begin{figure}[H]
\centering
\includegraphics[width=0.92\linewidth]{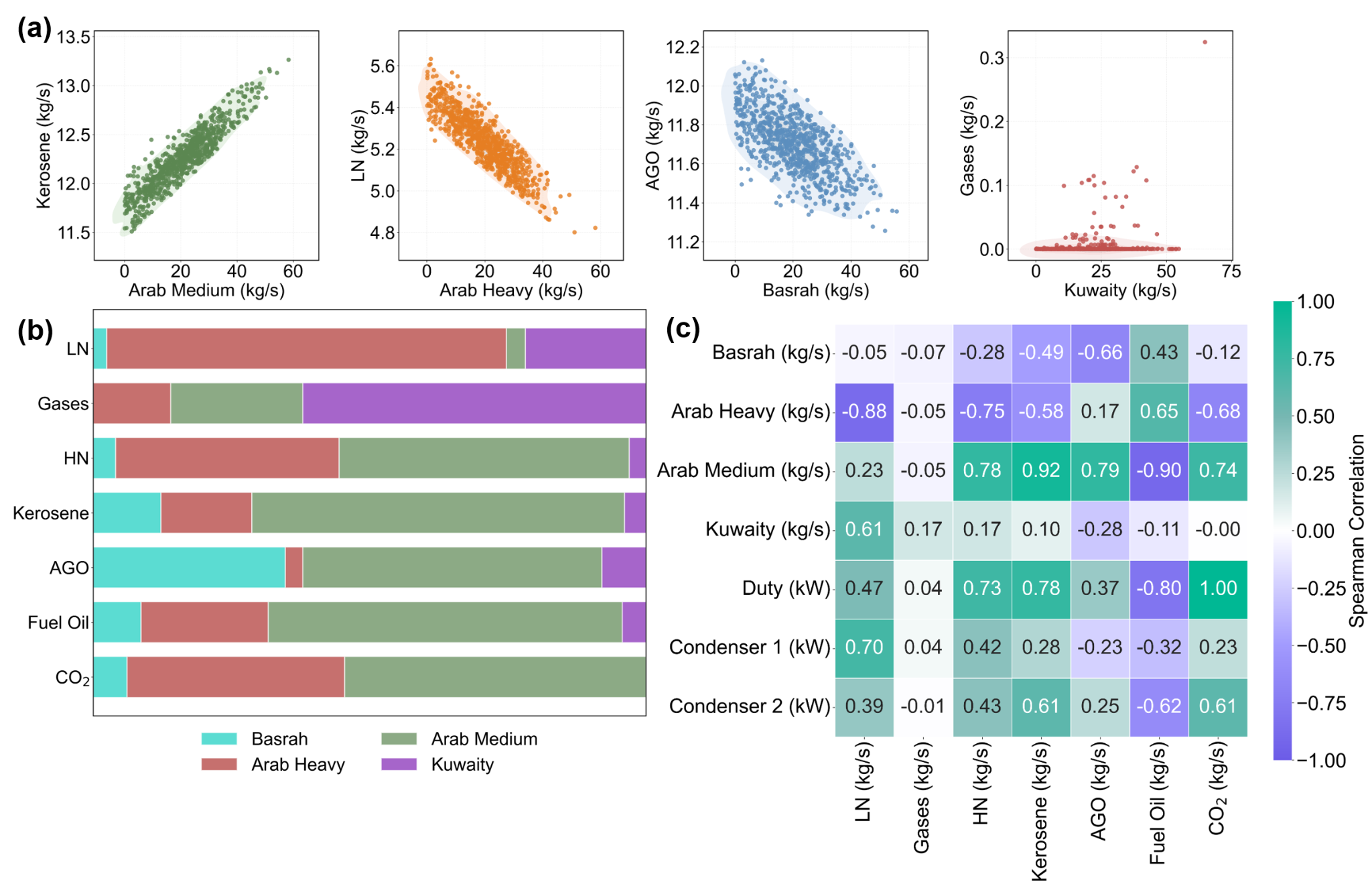}
\caption{Data characteristics for the crude distillation unit. Panel (a) shows kerosene increasing with Arab Medium flow while Light Naphtha decreases with Arab Heavy flow. Panel (b) shows relative crude contributions to product streams. Panel (c) shows the Spearman correlation map of the input and output variables.}
\label{fig_refinery_relationships}
\end{figure}

Figure~\ref{fig_refinery_relationships}(a) presents four representative bivariate relations. Kerosene rises with Arab Medium flow, while Light Naphtha falls as Arab Heavy increases. The corresponding Spearman correlations in Figure~\ref{fig_refinery_relationships}(c) are 0.92 and $-0.88$ for the two pairs respectively. Figure~\ref{fig_refinery_relationships}(c) also associates Arab Medium with Fuel Oil at $-0.90$ and Arab Heavy with Fuel Oil at 0.65. Meanwhile, Figure~\ref{fig_refinery_relationships}(b) compares the relative crude contributions represented in each product stream. Figure~\ref{fig_refinery_relationships}(c) further indicates that reboiler duty is positively associated with Kerosene at 0.78 and CO$_2$ at 1.00, while its correlation with Fuel Oil is $-0.80$.

\subsubsection{Surrogate performance and residual screening}

\begin{figure}[H]
\centering
\begin{overpic}[width=0.96\linewidth]{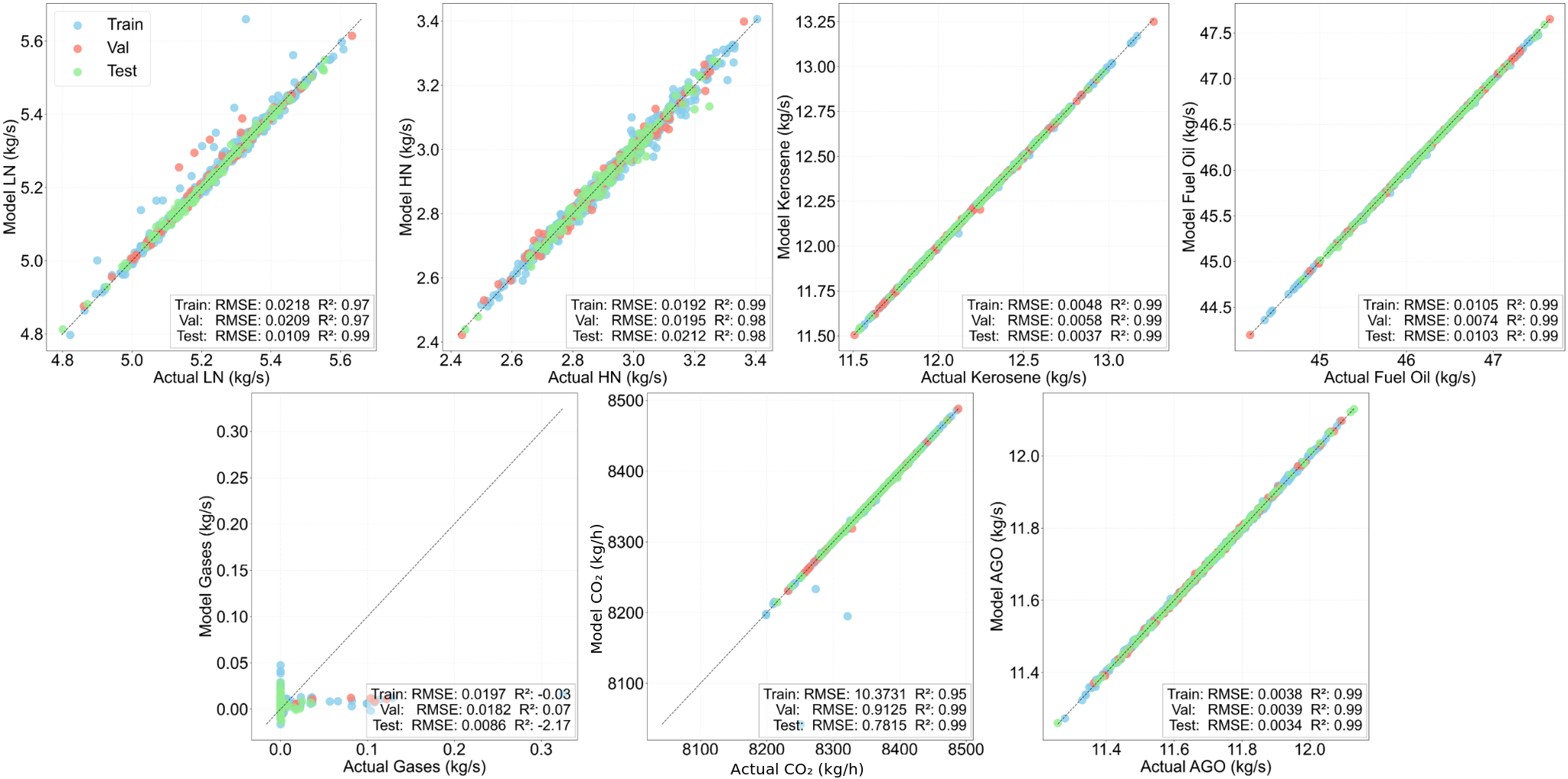}
\put(1,47){\colorbox{white}{\footnotesize\bfseries (a)}}
\put(26,47){\colorbox{white}{\footnotesize\bfseries (b)}}
\put(51,47){\colorbox{white}{\footnotesize\bfseries (c)}}
\put(76,47){\colorbox{white}{\footnotesize\bfseries (d)}}
\put(14,23){\colorbox{white}{\footnotesize\bfseries (e)}}
\put(40,23){\colorbox{white}{\footnotesize\bfseries (f)}}
\put(65,23){\colorbox{white}{\footnotesize\bfseries (g)}}
\end{overpic}
\caption{Base-model parity plots across training, validation, and test data for (a) Light Naphtha, (b) Heavy Naphtha, (c) Kerosene, (d) Fuel Oil, (e) Gases, (f) CO$_2$, and (g) AGO. The six non-gas surrogates provide reasonable simulator approximations within the sampled range.}
\label{fig_refinery_parity}
\end{figure}

Figure~\ref{fig_refinery_parity} presents the base-ANN parity plots on the training, validation, and test splits. Figure~\ref{fig_refinery_parity}(a)--(d), (f), and (g) place Light Naphtha, Heavy Naphtha, Kerosene, Fuel Oil, CO$_2$, and AGO close to the one-to-one line. Figure~\ref{fig_refinery_parity}(b) gives a test $R^2$ of 0.98 for Heavy Naphtha, while Figure~\ref{fig_refinery_parity}(a), (c), (d), (f), and (g) each give a value greater than 0.99. The corresponding test RMSE values are 21.2 g/s for Heavy Naphtha in Figure~\ref{fig_refinery_parity}(b), 10.3 g/s for Fuel Oil in Figure~\ref{fig_refinery_parity}(d), and 0.78 kg/h for CO$_2$ in Figure~\ref{fig_refinery_parity}(f). The six non-gas prediction errors are small relative to the observed response ranges. In contrast, Figure~\ref{fig_refinery_parity}(e) shows a poor fit for the gas stream. For the gas stream, the zero median and occasional larger flows make a small RMSE insufficient evidence of a useful fit. The test $R^2$ of $-2.17$ in Figure~\ref{fig_refinery_parity}(e) confirms that the model does not explain the variation among the less frequent non-zero observations and is less informative than the test-mean baseline.

Table~\ref{tab_refinery_models} shows that the effect of residual correction depends on the output. Corrections make only small changes to Heavy Naphtha, Kerosene, and AGO, but increase Light Naphtha test RMSE from 10.9 to 18.7 g/s. The Gases test $R^2$ remains negative at $-2.02$, despite residual correction. For CO$_2$, validation RMSE falls from 0.91 to 0.66 kg/h, while test RMSE rises from 0.78 to 2.55 kg/h. Fuel Oil shows the largest validation gain, with RMSE decreasing from 7.4 to 4.1 g/s, and its training RMSE also falls from 10.5 to 4.9 g/s. Fuel Oil is selected for the detailed reliability analysis under the rule in Equation~\ref{eq_validation_improvement}. Its test RMSE falls from 10.3 to 4.7 g/s, a reduction of about 54\%.

\begin{table}[H]
\centering
\caption{Base and residual-corrected performance for the crude distillation unit.}
\label{tab_refinery_models}
\scriptsize
\resizebox{\textwidth}{!}{%
\begin{tabular}{llcccccc}
\toprule
\multirow{2}{*}{\textbf{Target}} & \multirow{2}{*}{\textbf{Metric}} &
\multicolumn{2}{c}{\textbf{Training}} & \multicolumn{2}{c}{\textbf{Validation}} & \multicolumn{2}{c}{\textbf{Test}} \\
\cmidrule(lr){3-4}\cmidrule(lr){5-6}\cmidrule(lr){7-8}
& & \textbf{Base} & \textbf{Corrected} & \textbf{Base} & \textbf{Corrected} & \textbf{Base} & \textbf{Corrected} \\
\midrule
\multirow{2}{*}{Light Naphtha (kg/s)} & RMSE & 0.0218 & 0.0243 & 0.0209 & 0.0205 & 0.0109 & 0.0187 \\
& $R^2$ & 0.9745 & 0.9683 & 0.9734 & 0.9743 & 0.9945 & 0.9838 \\
\addlinespace
\multirow{2}{*}{Gases (kg/s)} & RMSE & 0.0197 & 0.0195 & 0.0182 & 0.0181 & 0.0086 & 0.0083 \\
& $R^2$ & $-0.0336$ & $-0.0171$ & 0.0702 & 0.0814 & $-2.1670$ & $-2.0193$ \\
\addlinespace
\multirow{2}{*}{Heavy Naphtha (kg/s)} & RMSE & 0.0192 & 0.0193 & 0.0195 & 0.0194 & 0.0212 & 0.0214 \\
& $R^2$ & 0.9860 & 0.9859 & 0.9850 & 0.9852 & 0.9819 & 0.9816 \\
\addlinespace
\multirow{2}{*}{Kerosene (kg/s)} & RMSE & 0.0048 & 0.0048 & 0.0058 & 0.0052 & 0.0037 & 0.0038 \\
& $R^2$ & 0.9997 & 0.9997 & 0.9996 & 0.9997 & 0.9998 & 0.9998 \\
\addlinespace
\multirow{2}{*}{AGO (kg/s)} & RMSE & 0.0038 & 0.0035 & 0.0039 & 0.0036 & 0.0034 & 0.0033 \\
& $R^2$ & 0.9994 & 0.9995 & 0.9993 & 0.9994 & 0.9996 & 0.9996 \\
\addlinespace
\multirow{2}{*}{Fuel Oil (kg/s)} & RMSE & 0.0105 & 0.0049 & 0.0074 & 0.0041 & 0.0103 & 0.0047 \\
& $R^2$ & 0.9997 & 0.9999 & 0.9998 & 0.9999 & 0.9997 & 0.9999 \\
\addlinespace
\multirow{2}{*}{CO$_2$ (kg/h)} & RMSE & 10.3731 & 6.5386 & 0.9125 & 0.6567 & 0.7815 & 2.5496 \\
& $R^2$ & 0.9503 & 0.9802 & 0.9996 & 0.9998 & 0.9997 & 0.9966 \\
\bottomrule
\end{tabular}}
\end{table}

\paragraph{Fuel Oil residual reliability diagnostic}

\begin{figure}[H]
\centering
\includegraphics[width=0.92\linewidth]{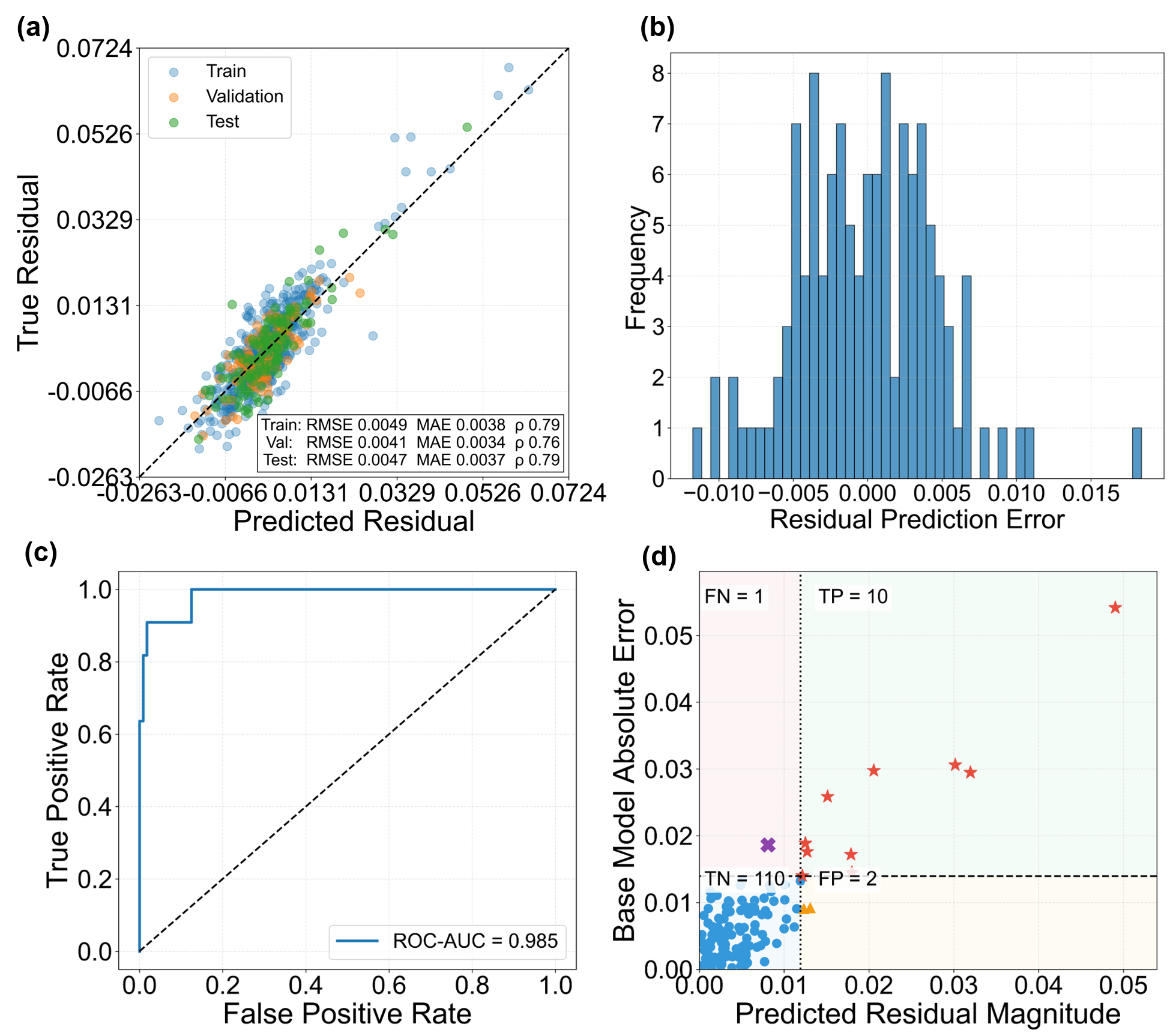}
\caption{Fuel Oil residual and reliability-screening analysis showing (a) predicted and true signed residuals, (b) 50-bin test correction-error distribution, (c) validation ROC curve, and (d) residual-score classification profile on the test split with the locked threshold.}
\label{fig_refinery_weakness}
\end{figure}

The selected Fuel Oil surrogate is examined for local modelling strength (Table~\ref{tab_refinery_models}). Figure~\ref{fig_refinery_weakness}(a) compares predicted and true signed residuals. The RMSE and MAE values are 4.9 and 3.8 g/s for training, 4.1 and 3.4 g/s for validation, and 4.7 and 3.7 g/s for testing. The similar errors across the splits indicate that the correction model retains its point-prediction performance on unseen data. Figure~\ref{fig_refinery_weakness}(b) places most test correction errors near zero, although a small number of observations extend beyond 10 g/s in either direction. On the validation split, the absolute base error and the residual score have a Spearman correlation of 0.59. The calibration selects the 94th percentile of observed absolute base error and the 93rd percentile of residual score, which gives the 11.92 g/s residual-score threshold drawn in Figure~\ref{fig_refinery_weakness}(d). Figure~\ref{fig_refinery_weakness}(c) gives a validation AUC of 0.99. Figure~\ref{fig_refinery_weakness}(d) then applies the locked thresholds to the test split and gives 110 true negatives, 10 true positives, 2 false positives, and 1 false negative. These counts correspond to a precision of 0.83, recall of 0.91, and $F_1$ of 0.87. The score detects 10 of the 11 high-error Fuel Oil observations in the test split.

\FloatBarrier

\subsubsection{Optimisation Cases}

\paragraph{CO$_2$-constrained production study}

The refinery objective adds the normalised product predictions and subtracts the three scaled utility inputs. The optimisation problem is written as

\begin{equation}
\begin{aligned}
\max_x\quad
J_{\mathrm{ref}}(x)
&=
\sum_{p\in\mathcal{P}_{\mathrm{ref}}}
\frac{\hat{y}^{B}_p(x)-y_p^{\min}}
{y_p^{\max}-y_p^{\min}}
-x_{\mathrm{Duty}}
-x_{\mathrm{C1}}
-x_{\mathrm{C2}}\\
\text{subject to}\quad
80
&\leq
\sum_{j\in\mathcal{C}}\tilde{x}_j
\leq82\\
-2
&\leq
\sum_{j\in\mathcal{C}}\tilde{x}_j
-
\sum_{p\in\mathcal{P}_{\mathrm{ref}}}\hat{y}^{B}_p(x)
\leq2\\
y_p^{\mathrm{LB}}
&\leq
\hat{y}^{B}_p(x)
\leq
y_p^{\mathrm{UB}}
&&p\in\mathcal{P}_{\mathrm{ref}}\\
y_{\mathrm{CO_2}}^{\mathrm{LB}}
&\leq
\hat{y}^{B}_{\mathrm{CO_2}}(x)
\leq
y_{\mathrm{CO_2}}^{\mathrm{cap}}\\
D_M^2(\tilde{x})
&\leq
\delta_{\mathrm{ref}}^2\\
0
&\leq x_j\leq1
&&j\in\mathcal{X}.
\end{aligned}
\label{eq_refinery_optimisation}
\end{equation}

The refinery objective $J_{\mathrm{ref}}(x)$ is dimensionless. $y_p^{\min}$ and $y_p^{\max}$ are the limits used to normalise each product prediction. The terms $x_{\mathrm{Duty}}$, $x_{\mathrm{C1}}$, and $x_{\mathrm{C2}}$ are the scaled utility decisions. The set $\mathcal{C}$ contains the four crude feeds, and $\tilde{x}_j$ is the physical flow of feed $j$. The product set $\mathcal{P}_{\mathrm{ref}}$ contains Light Naphtha, Gases, Heavy Naphtha, Kerosene, AGO, and Fuel Oil. Throughput is held between 80 and 82 kg/s. The approximate balance allows a difference of at most 2 kg/s between total crude feed and predicted product flow. $y_{\mathrm{CO_2}}^{\mathrm{LB}}$ is the lower bound on predicted CO$_2$, $y_{\mathrm{CO_2}}^{\mathrm{cap}}$ is the selected CO$_2$ ceiling, and $\delta_{\mathrm{ref}}$ is the refinery Mahalanobis-distance limit. The objective assigns unit weights to the six scaled product predictions and three scaled utility penalties.

\begin{figure}[H]
\centering
\includegraphics[width=0.96\linewidth]{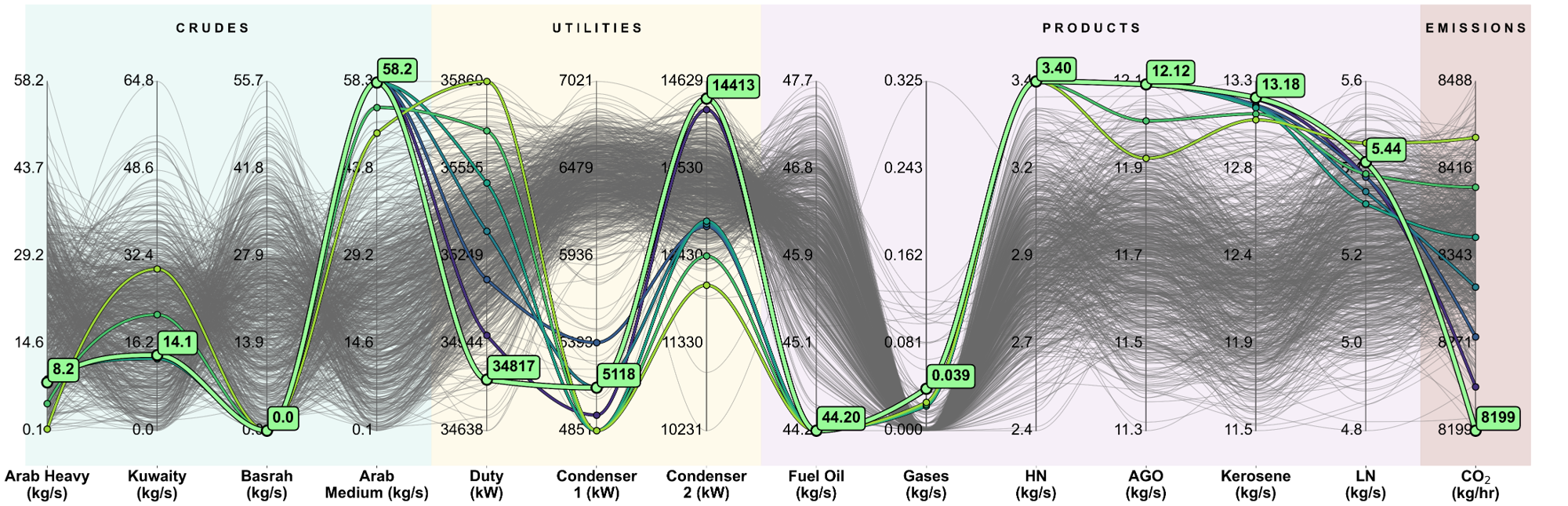}
\caption{Parallel-coordinate comparison of grey refinery simulation observations and coloured optimisation solutions under different CO$_2$ ceilings. The figure relates the crude blend and utility inputs to the predicted emissions and product rates for solutions that satisfy the imposed input-domain constraint.}
\label{fig_refinery_solutions}
\end{figure}

Figure~\ref{fig_refinery_solutions} is a parallel-coordinate comparison of the grey refinery simulation observations and the coloured solutions obtained under different CO$_2$ ceilings. It compares the joint changes in crude blend, utility inputs, predicted emissions, and product rates while the input-domain constraint is maintained. The numerical solution values and residual assessments are given in Table~\ref{tab_refinery_co2}.

All solutions in Table~\ref{tab_refinery_co2} exceed the locked Fuel Oil residual-score threshold of 11.92 g/s and are labelled WMR. The score is 13.1 g/s at the tightest CO$_2$ limit of 8,198.8 kg/h and rises to 101.5 g/s at 8,446.8 kg/h. The indicator half-width increases from 6.3 to 49.4 g/s over the same range. The first solution lies close to the threshold, whereas the remaining scores move progressively farther above it. These domain-compliant solutions have high Fuel Oil residual scores.

\begin{table}[H]
\centering
\caption{Fuel Oil model-region assessment for the refinery CO$_2$ sweep. The band is centred on corrected Fuel Oil, and its half-width is given in g/s. CO$_2$ is rounded to one decimal place and the remaining quantities to two.}
\label{tab_refinery_co2}
\scriptsize
\resizebox{\textwidth}{!}{%
\begin{tabular}{cccccc}
\toprule
\textbf{CO$_2$ limit (kg/h)} & \textbf{Base-model CO$_2$ (kg/h)} & \textbf{Corrected Fuel Oil (kg/s)} & \textbf{Residual score (g/s)} & \textbf{Band half-width (g/s)} & \textbf{Model region} \\
\midrule
8,198.8 & 8,198.8 & 44.21 & 13.10 & 6.30 & WMR \\
8,240.2 & 8,235.2 & 44.22 & 26.10 & 12.60 & WMR \\
8,281.5 & 8,276.5 & 44.23 & 35.10 & 17.50 & WMR \\
8,322.8 & 8,317.8 & 44.25 & 49.70 & 23.60 & WMR \\
8,364.2 & 8,359.2 & 44.26 & 65.70 & 32.90 & WMR \\
8,405.5 & 8,400.5 & 44.28 & 84.30 & 41.10 & WMR \\
8,446.8 & 8,441.8 & 44.30 & 101.50 & 49.40 & WMR \\
Unconstrained & 8,198.8 & 44.21 & 13.10 & 6.30 & WMR \\
\bottomrule
\end{tabular}}
\end{table}

\paragraph{Arab Heavy contract sensitivity}

In this case, the Arab Heavy operating level is held near a prescribed scaled target while the remaining variables are re-optimised. Mathematically, this constraint is written as

\begin{equation}
-0.01
\leq
x_{\mathrm{ArabHeavy}}-x_{\mathrm{ArabHeavy}}^{\mathrm{target}}
\leq
0.01.
\label{eq_heavy_target}
\end{equation}

Here, $x_{\mathrm{ArabHeavy}}$ is the optimised scaled Arab Heavy input and $x_{\mathrm{ArabHeavy}}^{\mathrm{target}}$ is its prescribed scaled target. Because the physical Arab Heavy range is 0.12 to 58.18 kg/s, a scaled tolerance of 0.01 corresponds to about 0.58 kg/s. This explains the difference between nominal and optimised physical values in Table~\ref{tab_heavy_sensitivity}.

\begin{figure}[H]
\centering
\includegraphics[width=0.80\linewidth]{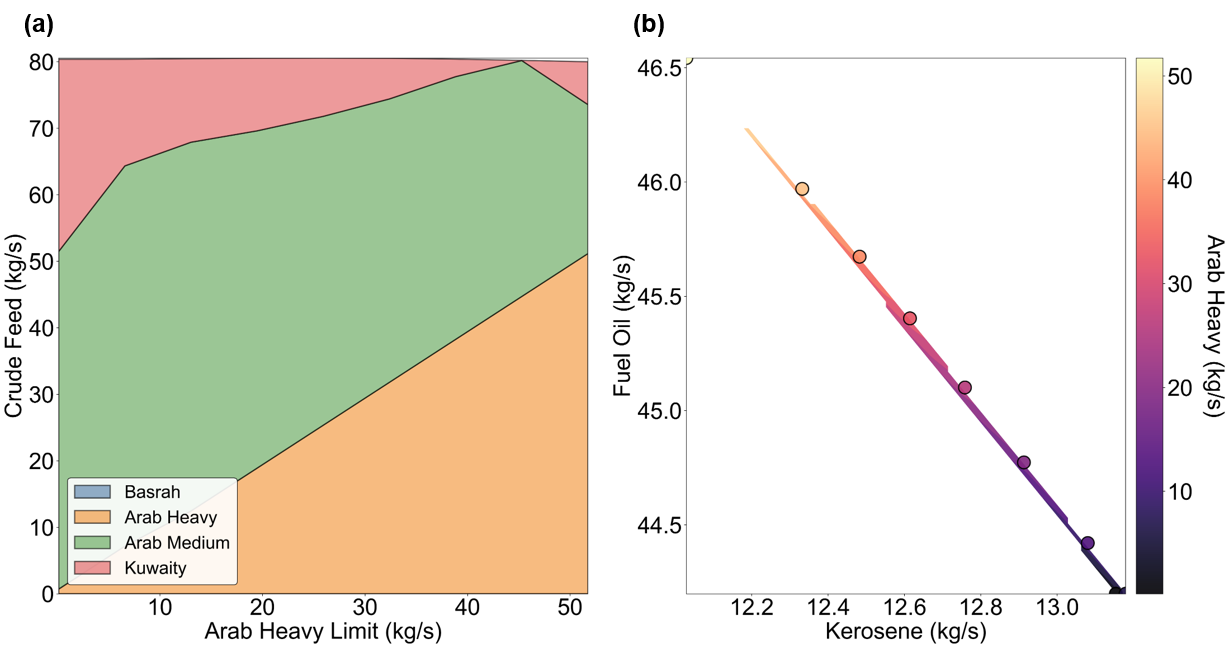}
\caption{Arab Heavy sensitivity results showing (a) re-optimised crude blend as the Arab Heavy target falls and (b) resulting changes in kerosene and corrected Fuel Oil predictions. Both panels show a predicted shift towards kerosene as Arab Heavy availability decreases.}
\label{fig_heavy_sensitivity}
\end{figure}

The contract sensitivity examines changes in the refinery predictions as Arab Heavy availability varies while the total crude-flow constraint is preserved. Figure~\ref{fig_heavy_sensitivity}(a) presents the re-optimised crude blend as the Arab Heavy target decreases from 51.73 to 0.12 kg/s. In Figure~\ref{fig_heavy_sensitivity}(a), Arab Medium and Kuwait provide most of the replacement feed, while Basrah remains comparatively small. Figure~\ref{fig_heavy_sensitivity}(b) relates the resulting Kerosene and corrected Fuel Oil predictions. In Figure~\ref{fig_heavy_sensitivity}(b), Kerosene rises from about 12.2 to 13.1 kg/s, whereas Fuel Oil falls from 46.55 to 44.22 kg/s. These values convey a predicted shift in the product slate towards Kerosene and away from Fuel Oil as Arab Heavy availability is reduced.

For Arab Heavy targets down to 19.48 kg/s, the solutions remain in GMR (Table~\ref{tab_heavy_sensitivity}). Lower targets move to WMR as Arab Medium and Kuwait replace more of the Arab Heavy feed under the same throughput constraint. The resulting change in product slate is accompanied by larger predicted Fuel Oil corrections. The score at the 13.02 kg/s target is only 0.38 g/s above $T_s$.

\begin{table}[H]
\centering
\caption{Fuel Oil model-region assessment for the Arab Heavy sensitivity. The band is centred on corrected Fuel Oil, and its half-width is given in g/s.}
\label{tab_heavy_sensitivity}
\scriptsize
\resizebox{\textwidth}{!}{%
\begin{tabular}{cccccc}
\toprule
\textbf{Arab Heavy target (kg/s)} & \textbf{Optimised Arab Heavy (kg/s)} & \textbf{Corrected Fuel Oil (kg/s)} & \textbf{Residual score (g/s)} & \textbf{Band half-width (g/s)} & \textbf{Model region} \\
\midrule
51.73 & 51.15 & 46.55 & 9.20 & 4.50 & GMR \\
45.28 & 44.69 & 45.98 & 11.00 & 5.30 & GMR \\
38.83 & 38.24 & 45.68 & 10.40 & 5.20 & GMR \\
32.38 & 31.79 & 45.41 & 7.30 & 3.50 & GMR \\
25.93 & 25.34 & 45.11 & 6.00 & 3.00 & GMR \\
19.48 & 18.89 & 44.78 & 8.70 & 4.20 & GMR \\
13.02 & 12.44 & 44.43 & 12.30 & 6.00 & WMR \\
6.57 & 7.16 & 44.21 & 14.60 & 7.00 & WMR \\
0.12 & 0.71 & 44.22 & 23.70 & 11.90 & WMR \\
\bottomrule
\end{tabular}}
\end{table}

\FloatBarrier

\subsection{Case Study 2. 395 MW Industrial Gas Turbine System}

\subsubsection{System and data description}
The plant study by Ashraf and Dua \cite{AshrafDua2024} provides records from a 395 MW gas turbine unit for the industrial case. Ambient air is compressed before entering the combustion system. Preheated natural gas is mixed with the compressed air and undergoes combustion. The hot combustion products expand through the gas turbine to drive the electrical generator. The exhaust supplies a heat recovery steam generator for the associated steam cycle. The present analysis models the gas turbine at unit level through the variables that describe its air path, fuel path, ambient conditions, and performance.

The selection of input variables follows the source study \cite{AshrafDua2024}, in which plant process and performance engineers identified variables used during power ramp-up and ramp-down. Compressor discharge pressure (CDP) and compressor discharge temperature (CDT) represent the compressed-air state before combustion. Gas fuel flow rate (GFFR), performance-heater gas outlet temperature (PHGOT), and fuel gas temperature at the combustion-chamber inlet (FGT) describe the fuel-side operating state. Ambient temperature (AT), ambient pressure (AP), and ambient humidity (AH) represent the air conditions of the site. The three output variables to be modelled are unit electrical power generation (Power), thermal efficiency (TE), and turbine heat rate (THR). The descriptive statistics of the dataset \cite{AshrafDua2024} are provided in Table~\ref{tab_gt_statistics}. The input and output variables cover wide operating ranges, and Power varies from 185.7 to 395.7 MW. These ranges indicate that ramping operation of the gas turbine is represented in the data.

\begin{table}[H]
\centering
\caption{Descriptive statistics for the gas turbine dataset. All values use two decimal places.}
\label{tab_gt_statistics}
\small
\begin{tabular}{lccccc}
\toprule
\textbf{Variable} & \textbf{Minimum} & \textbf{Median} & \textbf{Maximum} & \textbf{Mean} & \textbf{SD} \\
\midrule
Compressor discharge temperature ($^\circ$F) & 813.00 & 847.80 & 925.80 & 861.00 & 34.20 \\
Compressor discharge pressure (psi) & 185.90 & 232.70 & 312.00 & 247.90 & 36.90 \\
Gas fuel flow rate (lb/s) & 28.56 & 37.81 & 49.83 & 39.20 & 5.67 \\
Performance-heater gas outlet temperature ($^\circ$F) & 399.50 & 411.20 & 424.90 & 411.20 & 2.50 \\
Fuel gas temperature at combustor inlet ($^\circ$F) & 484.00 & 517.60 & 534.90 & 513.50 & 14.80 \\
Ambient temperature ($^\circ$C) & 20.50 & 25.60 & 34.80 & 26.30 & 3.60 \\
Ambient pressure (hPa) & 982.70 & 988.70 & 991.80 & 988.10 & 2.00 \\
Ambient humidity (\%) & 34.20 & 67.60 & 98.00 & 66.40 & 14.10 \\
Thermal efficiency (\%) & 32.44 & 39.02 & 43.05 & 38.96 & 2.38 \\
Power (MW) & 185.70 & 280.30 & 395.70 & 296.40 & 59.50 \\
Turbine heat rate (kJ/kWh) & 8,363.00 & 9,227.00 & 11,099.00 & 9,275.00 & 585.00 \\
\bottomrule
\end{tabular}
\end{table}

\subsubsection{Base and corrected surrogate performance}

\begin{figure}[H]
\centering
\includegraphics[width=0.92\linewidth]{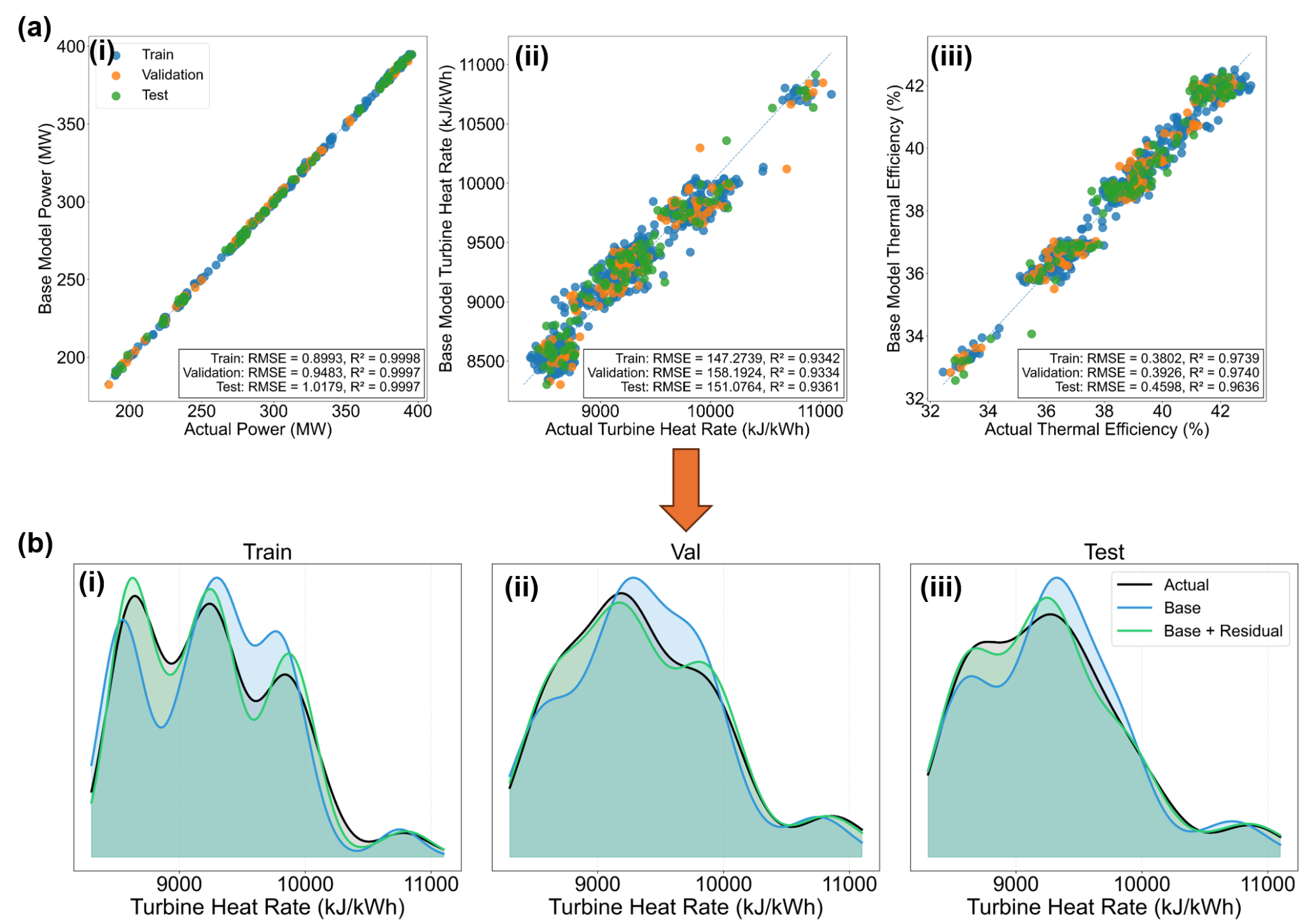}
\caption{Gas turbine model performance showing (a) base-model parity plots for Power, THR, and TE, alongside (b) density curves of measured, base, and corrected THR on the training, validation, and test splits.}
\label{fig_gt_models}
\end{figure}

Residual correction is evaluated separately for Power, TE, and THR. Figure~\ref{fig_gt_models}(a)(i) places the Power predictions close to the one-to-one line, with a test RMSE of 1.02 MW over the observed range of 185.7 to 395.7 MW and $R^2>0.99$. Figure~\ref{fig_gt_models}(a)(iii) gives a TE test RMSE of 0.46 percentage points and $R^2=0.96$. In contrast, Figure~\ref{fig_gt_models}(a)(ii) shows wider scatter for THR, particularly at higher values, with a test RMSE of 151 kJ/kWh and $R^2=0.94$. Figure~\ref{fig_gt_models}(b)(i)--(iii) compares measured, base, and corrected THR distributions on the training, validation, and test splits. Residual correction lowers THR error across all three splits, with the largest relative validation gain among the gas turbine outputs. Test RMSE falls from 151 to 105 kJ/kWh.

The effect of correction differs among the gas turbine outputs (Table~\ref{tab_gt_models}). Power test RMSE increases from 1.02 to 1.08 MW, while TE test RMSE remains at 0.46 percentage points. THR validation RMSE decreases from 158 to 99 kJ/kWh, which is a 37\% improvement. Its test RMSE decreases from 151 to 105 kJ/kWh and test $R^2$ rises from 0.94 to 0.97. THR has the largest validation improvement and is selected for the detailed reliability analysis.

\begin{table}[H]
\centering
\caption{Gas turbine base and residual-corrected performance.}
\label{tab_gt_models}
\scriptsize
\resizebox{\textwidth}{!}{%
\begin{tabular}{llcccccc}
\toprule
\multirow{2}{*}{\textbf{Target}} & \multirow{2}{*}{\textbf{Metric}} &
\multicolumn{2}{c}{\textbf{Training}} & \multicolumn{2}{c}{\textbf{Validation}} & \multicolumn{2}{c}{\textbf{Test}} \\
\cmidrule(lr){3-4}\cmidrule(lr){5-6}\cmidrule(lr){7-8}
& & \textbf{Base} & \textbf{Corrected} & \textbf{Base} & \textbf{Corrected} & \textbf{Base} & \textbf{Corrected} \\
\midrule
\multirow{2}{*}{Power (MW)} & RMSE & 0.90 & 0.79 & 0.95 & 0.84 & 1.02 & 1.08 \\
& $R^2$ & 0.9998 & 0.9998 & 0.9997 & 0.9998 & 0.9997 & 0.9997 \\
\addlinespace
\multirow{2}{*}{THR (kJ/kWh)} & RMSE & 147.27 & 88.00 & 158.19 & 99.02 & 151.08 & 104.86 \\
& $R^2$ & 0.9342 & 0.9765 & 0.9334 & 0.9739 & 0.9361 & 0.9692 \\
\addlinespace
\multirow{2}{*}{Thermal efficiency (\%)} & RMSE & 0.38 & 0.38 & 0.39 & 0.39 & 0.46 & 0.46 \\
& $R^2$ & 0.9739 & 0.9738 & 0.9740 & 0.9743 & 0.9636 & 0.9638 \\
\bottomrule
\end{tabular}}
\end{table}

\subsubsection{THR residual reliability diagnostic}

\begin{figure}[H]
\centering
\includegraphics[width=0.92\linewidth]{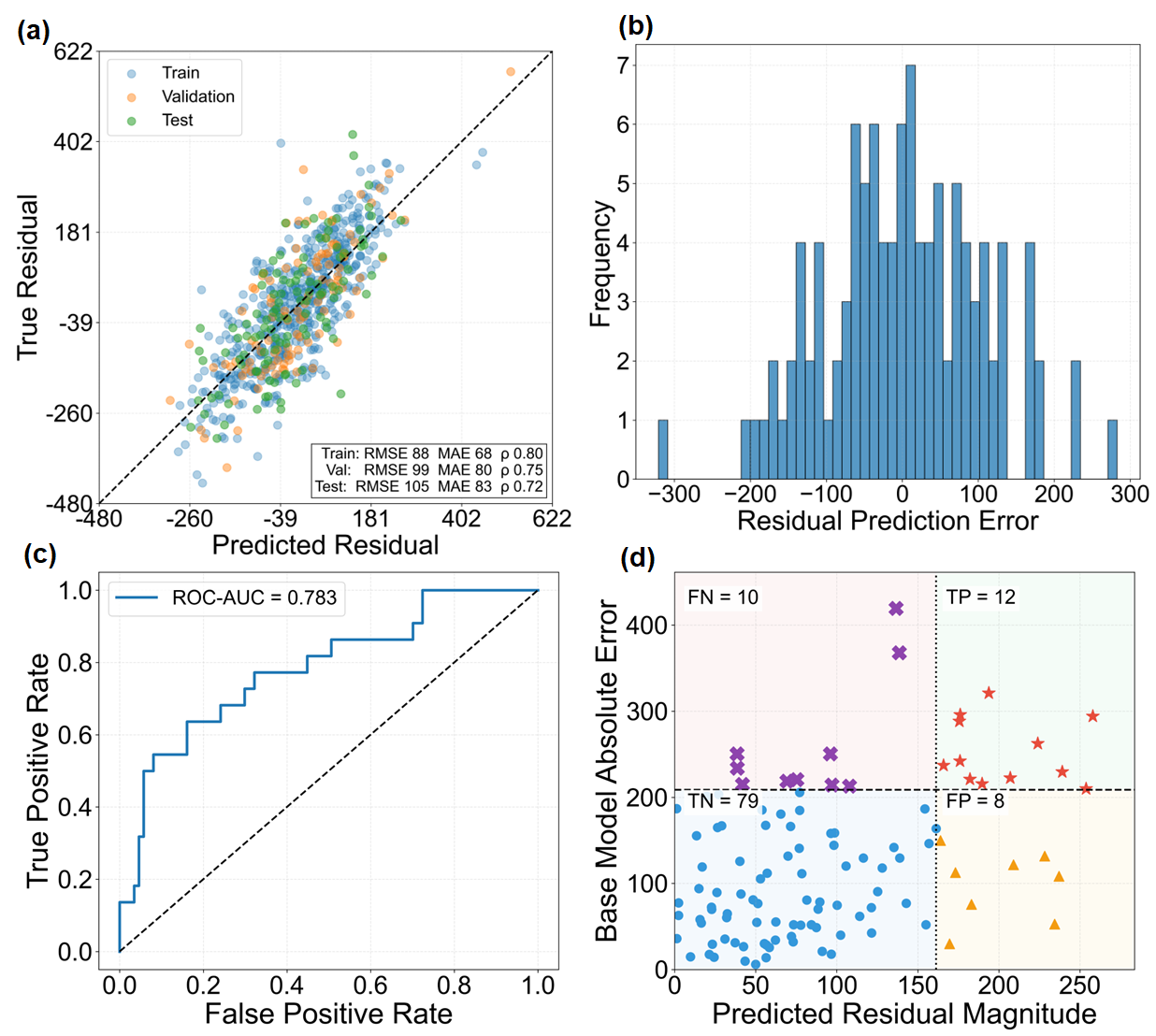}
\caption{THR residual and reliability-screening analysis showing (a) predicted and true signed residuals, (b) test correction-error distribution, (c) validation ROC curve, and (d) residual-score density and test classification with the locked threshold.}
\label{fig_gt_weakness}
\end{figure}

THR residual-score screening is evaluated against the high-error reference labels. Figure~\ref{fig_gt_weakness}(a) compares predicted and true signed residuals and gives Spearman correlations of 0.80 for training, 0.75 for validation, and 0.72 for testing. Figure~\ref{fig_gt_weakness}(a) also reports residual RMSE values of 88, 99, and 105 kJ/kWh and MAE values of 68, 80, and 83 kJ/kWh across the same splits. These values indicate that the model captures much of the direction and ordering of the required correction. The test correction-error distribution is centred near zero but extends from about $-300$ to $+300$ kJ/kWh (Figure~\ref{fig_gt_weakness}(b)). Residual correction lowers average THR error, although some individual corrections remain inaccurate. Screening uses the residual magnitude rather than its sign. On the validation split, the Spearman correlation between absolute base error and residual score is 0.35. The calibration selects an observed-error threshold of 208.62 kJ/kWh and a residual-score threshold of 161.32 kJ/kWh, and Figure~\ref{fig_gt_weakness}(c) gives a validation AUC of 0.78. Figure~\ref{fig_gt_weakness}(d) applies these fixed thresholds to the test split and gives 79 true negatives, 12 true positives, 8 false positives, and 10 false negatives. These counts correspond to a precision of 0.60, recall of 0.55, and $F_1$ of 0.57, which indicates moderate discrimination.

\subsubsection{Thermal-efficiency optimisation across fuel-flow targets}

The gas turbine study maximises the base-model TE prediction while keeping the three base outputs within their observed ranges. The Mahalanobis limit is set to the 95th percentile of the reference distances. A scaled GFFR target is imposed to examine the solution along the fuel-flow range. The optimisation problem for the gas turbine is written as

\begin{equation}
\begin{aligned}
\max_{x}\quad
&\hat{y}^{B}_{\mathrm{TE}}(x)\\
\text{subject to}\quad
&y_q^{\mathrm{LB}}
\leq
\hat{y}^{B}_q(x)
\leq
y_q^{\mathrm{UB}}
&&q\in\{\mathrm{TE},\mathrm{Power},\mathrm{THR}\}\\
&D_M^2(\tilde{x})
\leq
\delta_{95}^2\\
&0\leq x_j\leq1
&&j\in\mathcal{X}\\
&-0.01
\leq
x_{\mathrm{GFFR}}-x_{\mathrm{GFFR}}^{\mathrm{target}}
\leq
0.01.
\end{aligned}
\label{eq_gt_optimisation}
\end{equation}

In Equation~\ref{eq_gt_optimisation}, $\hat{y}^{B}_{\mathrm{TE}}(x)$ is the base-ANN prediction of thermal efficiency and $q$ indexes the constrained outputs TE, Power, and THR. $y_q^{\mathrm{LB}}$ and $y_q^{\mathrm{UB}}$ are the corresponding output bounds. $\delta_{95}$ is the 95th-percentile Mahalanobis-distance limit, and $x_{\mathrm{GFFR}}^{\mathrm{target}}$ is the prescribed scaled GFFR target around which $x_{\mathrm{GFFR}}$ may vary by 0.01. The nominal scaled targets are 0.2, 0.4, 0.6, 0.8, and 1.0. Using the observed range of 28.56 to 49.83 lb/s, the operating levels of GFFR correspond to nominal physical targets of 32.81, 37.07, 41.32, 45.58, and 49.83 lb/s. The scaled tolerance of 0.01 permits an adjustment of about 0.21 lb/s. IPOPT solves each case through the GAMS solver interface, after which the THR residual model supplies the signed correction, its magnitude as the score, and the indicator band.

\begin{table}[H]
\centering
\caption{Gas turbine optimisation and post-optimisation THR assessment. The THR indicator band is centred on the corrected THR prediction.}
\label{tab_gt_optimisation}
\scriptsize
\resizebox{\textwidth}{!}{%
\begin{tabular}{cccccc}
\toprule
\textbf{Scaled GFFR} & \textbf{Nominal GFFR (lb/s)} & \textbf{Base-model TE (\%)} & \textbf{Signed THR residual (kJ/kWh)} & \textbf{THR indicator band (kJ/kWh)} & \textbf{Model region} \\
\midrule
0.2 & 32.81 & 38.74 & $-462.31$ & $9,208.45\pm224.24$ & WMR \\
0.4 & 37.07 & 41.30 & $-424.97$ & $8,727.05\pm206.14$ & WMR \\
0.6 & 41.32 & 42.94 & $-229.65$ & $8,432.24\pm114.46$ & WMR \\
0.8 & 45.58 & 43.05 & $-106.29$ & $8,480.86\pm50.38$ & GMR \\
1.0 & 49.83 & 42.44 & 163.06 & $8,525.73\pm81.53$ & WMR \\
\bottomrule
\end{tabular}}
\end{table}

Predicted TE reaches the observed upper bound of 43.05\% at the 0.8 GFFR target (Table~\ref{tab_gt_statistics}), then falls at the 1.0 target. At the 0.8 target, the TE bound is active and the THR residual score gives the only GMR label. The large negative THR corrections at the three lower GFFR targets shift to a positive correction at the highest target, whose score is only 1.74 kJ/kWh above $T_s$ (Table~\ref{tab_gt_optimisation}).

\FloatBarrier

\section{Conclusion}
\label{sec_conclusion}

ML surrogates can be accurate on their aggregated scores while their predictive accuracy varies across the local input space. The variation in local predictive accuracy limits their use in the decision-making for operation of safety-critical thermal systems. This paper developed the TSRL framework, which attaches a point-specific modelling-strength label (good or weak) to each computed optimal solution, facilitating the decision-making for the operation of thermal systems. As a case study, The TSRL framework is applied to a simulated crude distillation unit and a 395 MW industrial gas turbine and the strength of computed optimal solutions under various operating scenarios is investigated.

Residual correction improved Fuel Oil and THR predictions, although its effect depended on the output. Screening was more effective for Fuel Oil than for THR. All eight solutions of the refinery CO$_2$ sweep satisfied the Mahalanobis constraint but were labelled WMR, and the Arab Heavy sensitivity moved to WMR at the three lowest targets. For the gas turbine, the scaled GFFR target of 0.8 gave the highest predicted TE and the only GMR label among the five targets.

The domain constraints keep the solution search close to the data envelopes, while the residual score distinguishes GMR and WMR within the data-driven domain. The optimal solutions computed through TSRL ensures input-domain compliance and providing the information about the local modelling strength which facilitates the decision-making for the operation of thermal systems.

\section*{Limitations and future work}

The thresholds are calibrated for one simulated refinery dataset and one gas turbine unit, which limits their transfer across operating regimes. Transfer to another unit will require checking score calibration against its operating data and, where available, an engineering error limit. The present GMR and WMR labels remain relative to validation error percentiles. A single random split was used for each case, leaving sensitivity to alternative partitions and temporal drift unquantified. Repeated and chronological partitions could assess this sensitivity as additional industrial units and operating regimes are tested. Screened optimisation solutions also require independent checks through simulator reruns or controlled plant trials.

\section*{Author statement}
\textbf{Talha Ansar:} Methodology, Formal Analysis, Writing – Original Draft,  \textbf{Waqar Muhammad Ashraf:} Conceptualization, Validation, Data Curation, Methodology,  Writing – Review \& Editing, \textbf{Abdulelah S. Alshehri:} Methodology,  Writing – Review \& Editing, Funding \textbf{Omar Bayomie:} Conceptualization, Methodology, Validation, Data Curation, Software, Writing – Review \& Editing, Project Supervision.

\section*{Acknowledgements}
The authors acknowledge that they have used LLMs (Claude - Opus) for improving grammar and enhancing the readability of manuscript. They have thoroughly reviewed the article and take full responsibility of the work. Abdulelah S. Alshehri would like extend his appreciation to King Saud University for funding his research through the Ongoing Research Funding program, (ORF-2026-1455), King Saud University, Riyadh, Saudi Arabia.

\section*{Declaration of competing interests}
The authors declare no competing financial or non-financial interests.

\bibliographystyle{unsrt}
\bibliography{references}

\end{document}